\documentclass{article}

\usepackage{PRIMEarxiv}

\usepackage[utf8]{inputenc} 
\usepackage[T1]{fontenc}    
\usepackage{hyperref}       
\usepackage{url}            
\usepackage{booktabs}       
\usepackage{amsfonts}       
\usepackage{nicefrac}       
\usepackage{microtype}      
\usepackage{lipsum}
\usepackage{fancyhdr}       
\usepackage{graphicx}       
\graphicspath{{media/}}     
\usepackage{makecell}
\usepackage{amsmath}
\usepackage{adjustbox}
\title{{Multi-domain learning CNN model for microscopy image classification}
}

\author{
  Duc Hoa Tran \\
  Department of Computer and Software Engineering, \\
  Polytechnique Montréal\\
  \texttt{duc-hoa.tran@polymtl.ca} \\
   \And
  Michel Meunier\\
  Department of Engineering Physics, \\
  Polytechnique Montréal \\
    \texttt{michel.meunier@polymtl.ca} \\
   \And
  Farida Cheriet \\
  Department of Computer and Software Engineering, \\
  Polytechnique Montréal\\
        \texttt{farida.cheriet@polymtl.ca} \\
}

\begin{document}
\maketitle

\begin{abstract}
For any type of microscopy image, getting a deep learning model to work well requires considerable effort to select a suitable architecture and time to train it. As there is a wide range of microscopes and experimental setups, designing a single model that can apply to multiple imaging domains, instead of having multiple per-domain models, becomes more essential. This task is challenging and somehow overlooked in the literature. In this paper, we present a multi-domain learning architecture for the classification of microscopy images that differ significantly in types and contents. Unlike previous methods that are computationally intensive, we have developed a compact model, called Mobincep, by combining the simple but effective techniques of depth-wise separable convolution and the inception module. We also introduce a new optimization technique to regulate the latent feature space during training to improve the network’s performance. We evaluated our model on three different public datasets and compared its performance in single-domain and multiple-domain learning modes. The proposed classifier surpasses state-of-the-art results and is robust for limited labeled data. Moreover, it helps to eliminate the burden of designing a new network when switching to new experiments.
\end{abstract}


\section{Introduction}

There exists a wide range of microscopy assays to reveal complex properties of cellular structures (tissues, cells, or subcellular components) and each set of images produced in a laboratory typically forms a different visual domain. Although Deep Learning (DL) models could yield excellent classification performance, they are highly specialized to each domain \cite{RebuffiNIPS2017, RebuffiBV18}. 
At the same time, designing and training an appropriate deep model are relatively complex operations to carry out successfully, even for experienced scientists \cite{Gaudenz2011}. Thus, there is growing interest in developing a single model that can be deployed for various biomedical studies without adjusting its parameters. This is a challenging task as it requires deep learning models to learn a unified feature representation across different domains.

To contribute to this research effort, we aimed at designing a deep convolution neural network (CNN) to learn unified representations for the classification of microscopy image sets that have significantly different characteristics. This problem belongs to Multi-Domain Learning (MDL) and can be distinguished from the related domain adaptation technique in two ways: the domain shift and the learning sequence. First, the domain shift refers to the visual difference between image domains, including image content (objects of interest) and image appearance (style). While standard domain adaptation methods deal with the change in style and not the objects of interest, our model handles both changes in image content and style. Second, in terms of the learning sequence, typical domain adaptation approaches learn multiple domains sequentially to maximize their performance in a target domain. However, after adapting from a source domain to the target one, the model cannot maintain its initial performance on the source domain or it cannot learn without forgetting \cite{RebuffiNIPS2017}. In this sense, domain adaptation is like transfer learning, where DL models are trained on a common large dataset, before being fine-tuned on the domain of interest. By contrast, our proposed model learns multiple domains simultaneously and aims at achieving high performance on all the learned domains.

In this work, we design a deep CNN architecture that combines an inception module and depth-wise separable convolution layers. These two techniques, introduced in GoogleNet \cite{GoogleNet2015} and \cite{Mobilenetv1_2017}, are popular in the design of many deep neural networks. However, to the best of our knowledge, this is the first work that explores their combination for multi-domain learning of microscopy images obtained from different imaging devices and with different objects of interest. The proposed model is lightweight and scalable. In addition, we introduce an optimization approach for feature regularization during training to enhance the network's performance, allowing it to beat the state-of-the-art models. 

To sum up, the major contributions of this study are:
\begin{itemize}
\item We propose an MDL model for learning simultaneously different microscopy image domains. It could work effectively in different applications without requiring the adaptation of domain-specific parameters.
\item We formulate a simple yet effective optimization function to regulate feature space, improving network performance. 
\item Our proposed model is remarkably compact and robust against limited available training data and outperforms the best results published as yet on three public datasets.
\end{itemize} 

\section{Related Works}
DL algorithms or deep neural networks have emerged as the dominant methods in every application of biomedical image analysis, including microscopy. Instead of using handcrafted feature extractors as in conventional machine learning methods, CNN models learn by themselves to extract the optimal features from input images. However, it is still challenging to design a model that can extract a unified feature representation from multiple microscopy image domains because of the highly specialized experiments. In the literature, most DL-based analyses have used transfer learning, as it can produce significantly better results than when training from scratch. Comprehensive reviews of different applications using transfer learning can be found in \cite{XingSurvey2018,MEIJERING2020}. In this approach, a DL model is pre-trained on a large dataset of labeled natural images like ImageNet and then fine-tuned on a target dataset that usually has a few labeled images.

DL-based domain adaptation approaches have also been investigated for digital pathology \cite{DomainAdapt2016, HuangZLDR17, GadermayrAKBM18}. In these approaches, the feature extractors of a DL model are first trained on a source domain and then adapted to the target domain via a retraining process. The two domains are supposed to be similar or undergo a minor domain shift, in the sense that the image style or appearance changes but the image content or objects of interest are the same. The domain adaptation methods conventionally tackle the problem by normalizing the imaging parameters, such as staining normalization, or aligning the source representation with the target one using feature space transforms \cite{Lafarge2019, SebagICLR2019}. Like in transfer learning, during fine-tuning of the DL network, the pre-trained parameters are specifically adjusted for the target domain and thus, the fine-tuned model cannot be reused on the source dataset. Many of the latest techniques use Generative Adversarial Networks (GAN) to learn domain-invariant features \cite{KoohbananiTMI2020}. For example, the authors of \cite{RenMICCAI18} propose to use adversarial training to align the feature distributions of shifted domains in the classification of prostate cancer images acquired from different scanners.

Recently, MDL has become a popular topic in computer vision, but there are still few published articles for biomedical image analysis, especially for microscopy images. The multi-domain adversarial learning approach presented in \cite{SebagICLR2019} was the first work using MDL in bio-image informatics. The authors experimented on cell-level fluorescence images obtained from three different centers and used a pre-trained VGG-16 \cite{VGG2015} network as the feature extractor to feed the cells classifier. Nevertheless, that study only considered the variation in image appearance, while the type of content (i.e cells) remained unchanged.
 
\section{Materials and Methods}

\subsection{Convolutional neural network architecture}
\label{section:CNN_arch}
In this section, we present a deep multi-domain CNN model, named Mobincep, which is compact but offers powerful classification capability for various types of microscopy images. To achieve efficient feature extraction, the network's design is based on the combination of the inception structure \cite{GoogleNet2015} and depth-wise separable convolution \cite{Mobilenetv1_2017}. Also, we describe a relevant training strategy, particularly the formulation of an integrated loss function for network optimization. The following sections describe the construction of our model.

\subsubsection{Depth-wise separable convolution layers}
\begin{figure}
	\centering
\includegraphics[width = 0.48\textwidth]{ 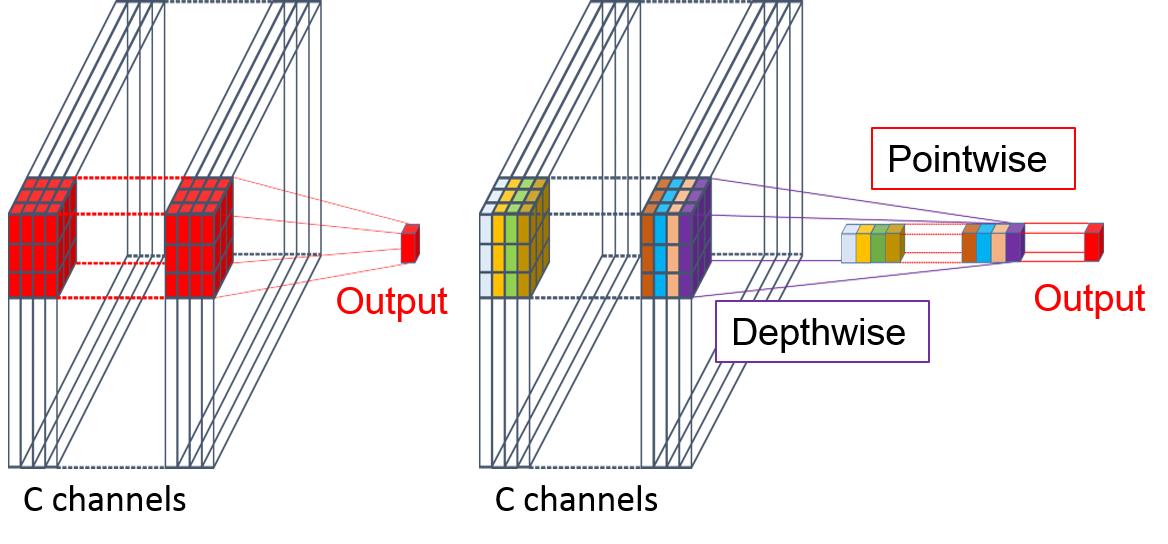}
	\caption{Example of depth-wise separable convolution compared with standard convolution operations. In the standard convolution (on the left), the multi-channel input is convolved with kernels having smaller spatial dimensions ($3 \times 3$) but the same
number of channels ($C$). The depth-wise separable convolution (on the right) divides the calculation into two separate and consecutive steps: depth-wise (convolution with $3 \times 3$ kernel of 1 channel) and point-wise (convolution with $1 \times 1$ kernel of $C$ channels).}
	\label{Fig:DepthwiseSepConv}
\end{figure}

The depth-wise separable convolution effectively reduces the computation complexity of the standard convolution by dividing the calculation into two separate and consecutive steps: depth-wise and point-wise (or $1 \times 1$) convolution \cite{GuoDepthwise19}. In the first step, each input channel is convolved with kernels that have only a single channel. Then, point-wise convolution creates a linear transformation of the corresponding output values across channels. We illustrate the two convolution approaches in Figure \ref{Fig:DepthwiseSepConv}. As depth-wise convolution reduces the number of deep CNN network parameters significantly, it also helps to decrease the possibility of over-fitting to a specific image dataset. 

Compared with standard convolution, the use of depth-wise separable convolution produces more discriminative features due to the decoupling of cross-channel and spatial correlations as suggested in \cite{Chollet17}. It can also help to promote the performance for learning natural images from multiple visual domains when replacing standard convolutions in a pre-trained ResNet-26 \cite{GuoDepthwise19}. Based on this observation, we exploit depth-wise separable convolution layers as one of the main strategies to design a compact model for microscopy images in multiple imaging settings.

\subsubsection{Inception module}
\begin{figure}
	\centering
		\includegraphics[width = 0.48\textwidth]{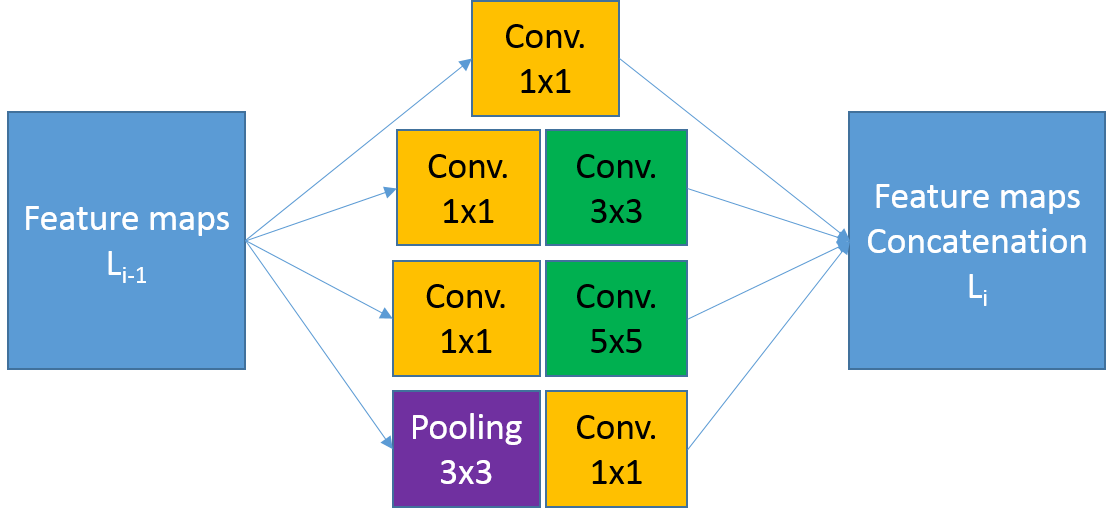}
	\caption{Design of the inception module. The feature maps of previous layer $L_{i-1}$ are processed by multiple branches with different reception fields. The outputs from each branch are concatenated into a set of feature maps $L_{i}$ to be used as the input to the following layer. }
	\label{Fig:InceptionModule}
\end{figure}

We illustrate the layout of an inception module in Figure \ref{Fig:InceptionModule}. It is a set of multiple convolution branches having different kernel sizes, where the output feature maps from each branch are concatenated and used as the input for the subsequent layer \cite{GoogleNet2015}. In this configuration, the use of the point-wise ($1 \times 1$) convolution and average pooling layer in each branch reduces the dimensions of the feature maps, leading to a remarkable reduction of multiplication operations. Meanwhile, the combination of various convolution kernel sizes helps to detect features at different scales. This is advantageous for microscopy images that typically express a wide variety of object morphologies and sizes. 

\subsubsection{Proposed network architecture}
\begin{figure*}
	\centering
		\includegraphics[height=0.4\textwidth, width = \textwidth]{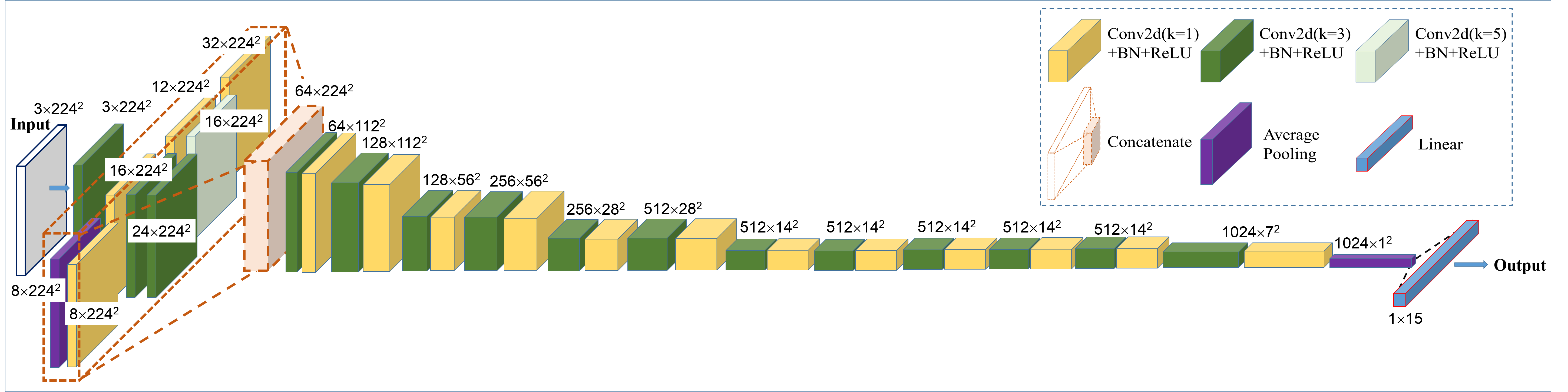}
\caption{The architecture of the proposed Mobincep model. Each of the convolution layers (Conv2d) is followed by a Batch Normalization layer (BN) and in-place Rectified Linear Unit (ReLU) activation function. The raw input images are passed through a buffer convolution layer to the inception module, which comprises four branches with different kernel sizes. Then, the concatenated feature maps are forwarded through a stack of multiple depth-wise separable convolution layers. After that, the extracted features are reduced in spatial size by average pooling operation and linearly combined to produce the output prediction.}
	\label{Fig:Mobincep}
\end{figure*}

The overall structure of our proposed Mobincep network is shown in Fig. \ref{Fig:Mobincep}. Through a buffer convolution layer, the raw input images are fetched to the inception module, comprising four branches with different kernel sizes. In this work, we choose $1 \times 1$, $3 \times 3$, and $5 \times 5$ filters as they have proved to be effective feature extractors. The concatenated feature maps are then delivered to a stack of multiple depth-wise separable convolution layers. Each of these layers is followed with Batch Normalization (BN) \cite{BatchNorm2015} and in-place Rectified Linear Unit (ReLU) \cite{ReLU2010} activation functions. The role of BN is to normalize each layer’s output such that it has zero mean and unit standard deviation, leading to faster convergence and bypassing local minima. For its part, the ReLU activation function has properties that help feed-forward neural networks to optimize easily with gradient-based convergence and generalize well on various data domains \cite{DeepLearningBook2016}. Lastly, the average pooling layer compresses the extracted feature maps into a feature vector. At the output, a linear layer combines the vector elements to produce the prediction probability for every image class. The class that has the highest probability from the output layer will be selected as the predicted class for the input image.

\subsection{Network optimization}
We propose a new model optimization approach to regularize the extracted feature space of microscopy images during training. The conventional loss function used in the optimization of deep CNN classifiers may not effectively regulate the latent space, leading to low performance and applicability to different sets of data. Therefore, we formulate two additional loss terms to encourage the feature representation of samples within each class to converge to a compact corresponding cluster. Assuming there are $K$ categories to be classified, the integrated cost function for training the Mobincep network is expressed as:
\begin{align}\label{eqn:cost_function}
L = L_{CE} + \gamma_1 \frac{1}{\Sigma_{k=1}^{K}(d^2(\mu_k,\mu))} + \gamma_2 \Sigma_{k=1}^{K}(s_k^2)
\end{align}
In equation (\ref{eqn:cost_function}), $L_{CE}$ is the conventional cross-entropy loss criterion for classification, which is calculated as:

\begin{align}\label{eqn:cross_entropy}
L_{CE} = -\frac{1}{N_B}\Sigma_{i=1}^{B} \Sigma_{c=1}^{K}I_{i,c} \log\frac{\exp(y_{i,c})}{\Sigma_{c=1}^{K}\exp(y_{i,c})} 
\end{align}
where $N_B$ is the training batch size; $I_{i,c} = 1 $ if label $c$ is the correct classification for image sample $i$ and $I_{i,c} = 0 $ otherwise; $y_{i,c}$ is the raw output probability of the network for the sample $i$ to have class label $c$. Also in equation \ref{eqn:cost_function}, we add the two other loss terms to impose an additional constraint for the training. In the first term, $d(\mu_k,\mu)$ denotes the distance between the centroid of each cluster in the feature space and the centroid of all latent points. In the second term, variable $s_k$ represents the scattering of each cluster, calculated as the sum of all distances between each latent point and the centroid $\mu_k$ of its cluster. Variables $\gamma_1$ and $ \gamma_2$ are the weights to balance the three loss terms in the total cost function.  
At each training iteration, the cluster centroids in the latent space are quickly determined by using the conventional K-Means clustering algorithm. 

Intuitively, when minimizing this loss function, the addition of two new loss terms helps in two ways: decrease the scattering (i.e. the embedded distances) of the input samples around their centroids and increase the embedded distance between clusters. This helps to better discriminate between clusters or classes.

\subsection{Experimental setup}
In this section, we describe our experiments with the Mobincep model on datasets from three different imaging domains.

\subsubsection{Datasets}
\label{sec:datasets}
\begin{figure}
	\centering	\includegraphics[width = 0.48\textwidth]{ 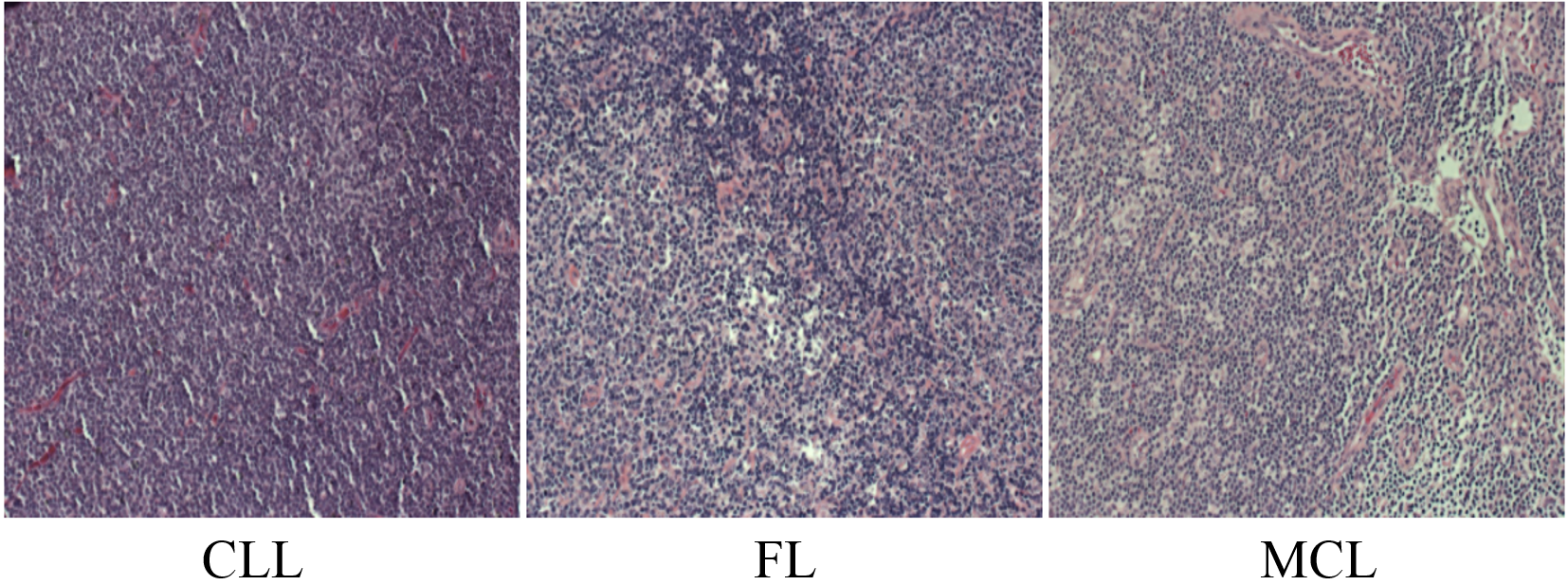}
\caption{Example of Lymphoma images in three different cases. The images were obtained from Hematoxylin- and Eosin- ($H\&E$) stained tissue samples using brightfield microscopy. There are three types of malignant lymphoma, i.e. cancer affecting lymph nodes: chronic lymphocytic leukemia (CLL), follicular lymphoma (FL), and mantle cell lymphoma (MCL).}
\label{Fig:lymphoma_imgs}

\end{figure}

\begin{figure}
	\centering	\includegraphics[width = 0.48\textwidth]{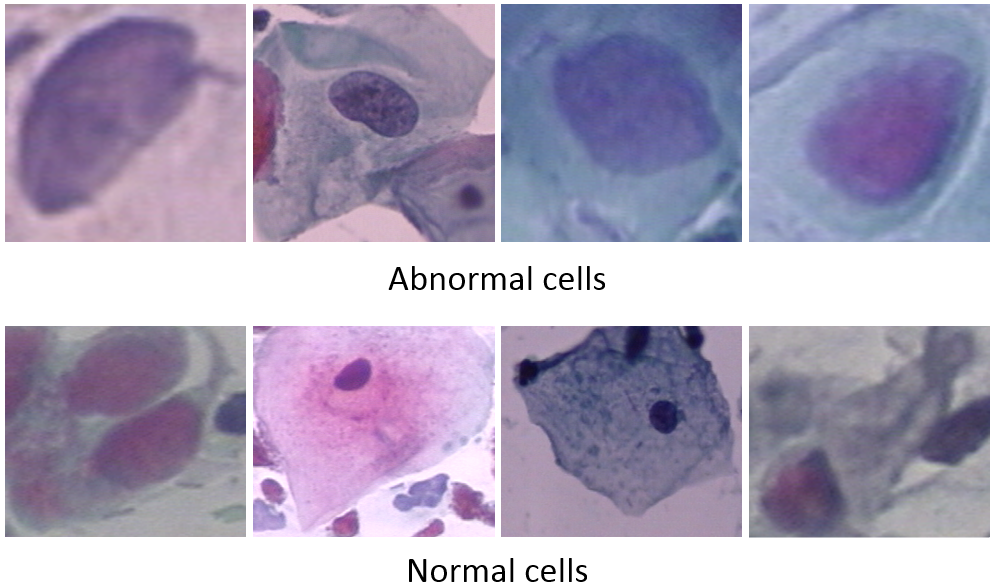}
\caption{Example of Pap-smear images in two cases: Abnormal  and Normal cells. In the sample preparation, a specimen of human cells is smeared onto a glass slide and colored using the Papanicolaou method. The abnormal cells are associated with the pre-cancerous stage.}
\label{Fig:Papsmear_imgs}
\end{figure}

\begin{figure}
	\centering	\includegraphics[width=0.48\textwidth]{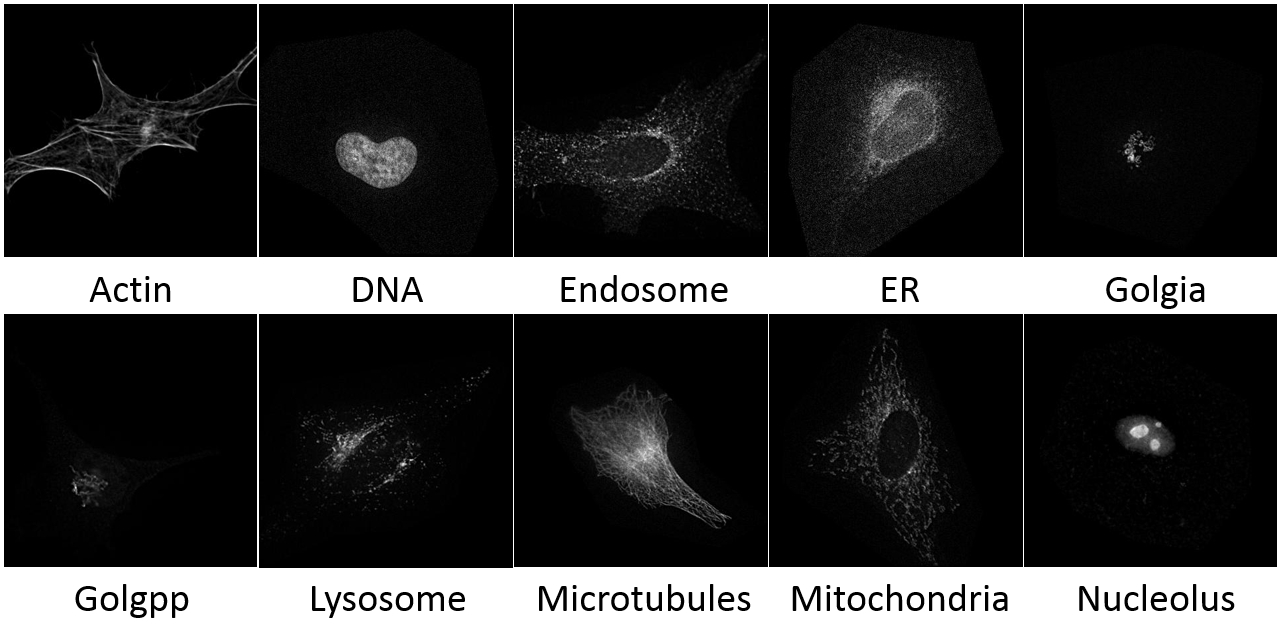}
\caption{Example images of the HeLa dataset. It comprises fluorescence microscopy images of sub-cellular organelles in HeLa cells, which are stained with various organelle-specific fluorescent dyes. There are ten categories: Actin, DNA(Nuclei), Endosomes, ER (Endoplasmic reticulum), Golgia (Giantin), Golgpp (GPP130), Lysosome, Microtubules, Mitochondria and Nucleolus. }
\label{Fig:hela_imgs}

\end{figure}

To create a dataset (Mix) that represents different imaging domains, we use three public microscopy datasets: Lymphoma (Lym), composed of tissue sample images \cite{IICBU2008}; Pap-smear (Pap), with images of cells \cite{Papsmear2005}; and HeLa, with images of sub-cellular organelles \cite{IICBU2008}. The characteristics of these datasets are summarized in Table \ref{table_dataset}. Example images from each dataset are shown in Figures \ref{Fig:lymphoma_imgs}, \ref{Fig:Papsmear_imgs} and \ref{Fig:hela_imgs}.

\begin{table}
\renewcommand{\arraystretch}{1.4}
\huge
\caption{Microscopy image datasets used experimentally} 
\label{table_dataset}
\centering
\begin{adjustbox}{width=0.48\textwidth}
\begin{tabular}{l|c|c|c}
\Xhline{5\arrayrulewidth}
\textbf{Dataset} & \# \textbf{Images} & \# \textbf{Classes}& \textbf{Dimensions}\\
\Xhline{3\arrayrulewidth}

Lymphoma & 375& 3& $1388\times1040\times3$\\
\hline
Pap-smear & 917& 2& \makecell{    $45\times43\times3$ \\ to $768\times284\times3$}\\
\hline
HeLa & 862& 10& $382\times382\times1$\\
\hline
Mix & 2154& 15&  \makecell{    $45\times43\times3$ \\ to $1388\times1040\times3$}\\
\Xhline{5\arrayrulewidth}
\end{tabular}
\end{adjustbox}
\end{table}

\subsubsection{Network training, validation and testing}
As shown in Table \ref{table_dataset}, the experimented datasets vary significantly in terms of image characteristics. For the model to handle them properly, we pre-processed the raw images by resizing them to $ 224 \times 224 \times 3$ and normalizing them to have the same dynamic intensity range. During training, we applied online data augmentation, combining a wide range of common image transformations, including rotation, flipping, cropping, and affine transformations (translate, scale, shear). We process the input images in batches of four. The network layers were initialized with the Kaiming uniform method \cite{KaimingHe2015}. We used the modified Adam optimizer, referred to as AMSGrad \cite{Adamamsgrad}, with a small learning rate of $10^{-4}$.
We validated the performance of the model using a 5-fold cross-validation strategy. In each fold, we randomly split each dataset into training, validation, and testing subsets. Specifically, we used 60$\%$ of the images for training, 20$\%$ for validation, and the remaining 20$\%$ to assess network performance. To obtain the model with the lowest validation error, the early stopping strategy was adopted, where the training stops when there is no better model after a certain number of training epochs (or patience period).
As the training process includes random processes (for instance, data augmentation), we ran the experiments five times and recorded the average result.
 
\subsection{Model complexity}
\begin{figure}
	\centering	\includegraphics[width = 0.48\textwidth]{ 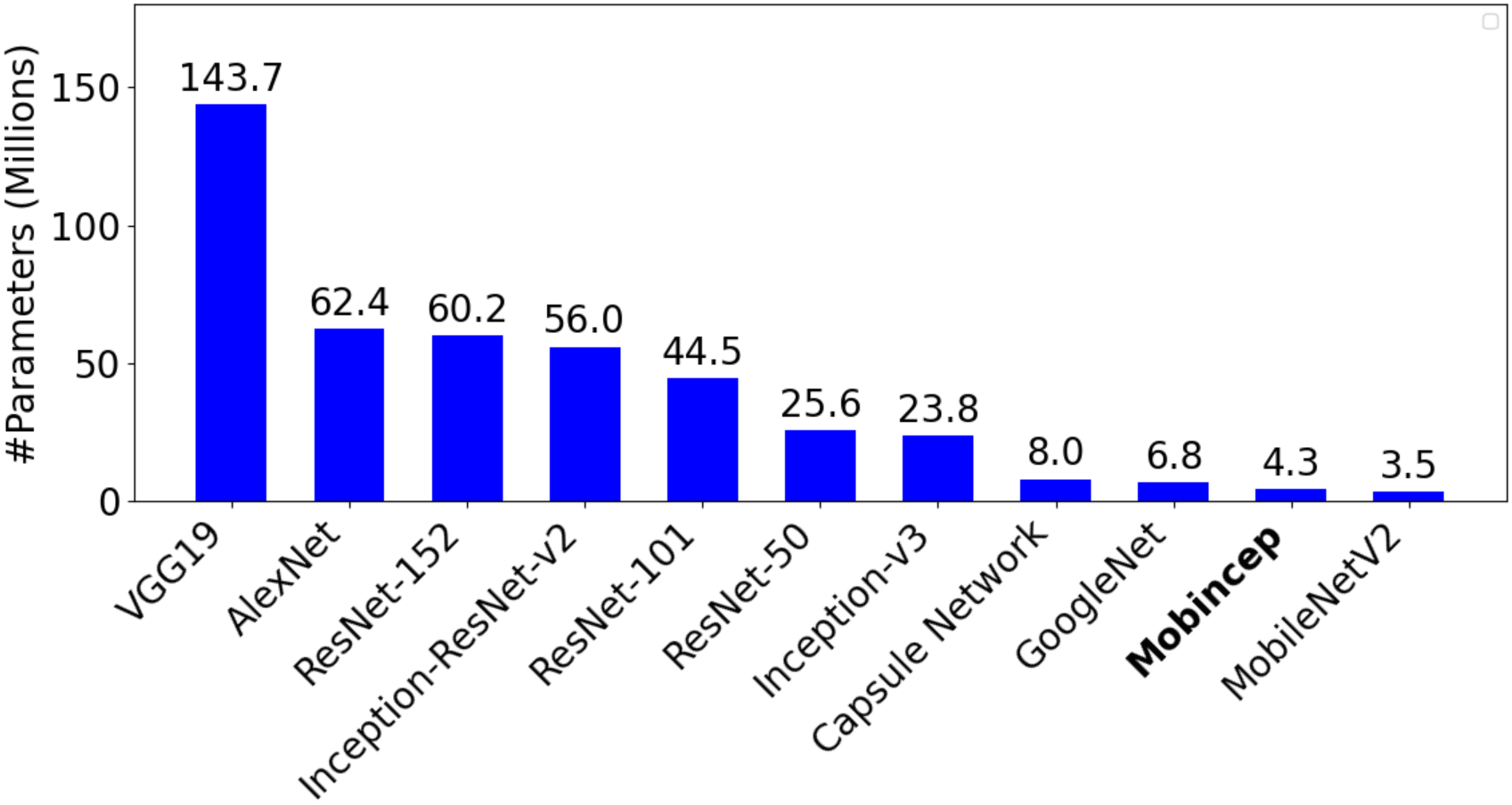}
\caption{Complexity of state-of-the-art CNN models compared by number of trainable parameters.}

\label{Fig:model_complexity}
\end{figure}

Figure \ref{Fig:model_complexity} shows the complexity of our proposed model compared to prominent CNN models from the literature. The number of trainable parameters in our Mobincep network is remarkably small compared with most of the recently published networks. For example, it requires about 13 times fewer parameters than Inception-ResNet-v2 \cite{SzegedyIV16}, which has 56 million parameters. Although the MobileNetV2 \cite{Mobilenetv22018} network has a slightly smaller number of parameters, our model can outperform MobileNetV2 by a large margin in various experiments, as shown in the following section.

It took less than 2 hours to train the model with each fold of the dataset. We used a GPU (model NVIDIA Tesla V100 SXM2 with 16 GB memory) to train the network in an end-to-end fashion. 

\section{Results}
We performed four main experiments to test the ability of the proposed model in dealing with microscopy images from three different visual domains. In each experiment, we report the results in both the MDL and Single-Domain Learning (SDL) modes.

\renewcommand{\arraystretch}{1.6}
\begin{table}[hbt!]
\centering
\huge
\caption{Top-1 classification accuracy on the different datasets compared to recent deep CNN models.}\label{table_SOA_1}
\begin{adjustbox}{width=0.48\textwidth}
\begin{tabular}{l|c|c|c|c}
\Xhline{5\arrayrulewidth}
\textbf{Model} & \textbf{Mix} & \textbf{Pap} & \textbf{Lym} & \textbf{HeLa} \\
\Xhline{2\arrayrulewidth}
VGG19 &83.92$\pm$2.1  & 84.32$\pm$1.9 & 83.52$\pm$1.7  & 84.73$\pm$2.1 \\
\hline
GoogleNet &84.11$\pm$2.3  & 84.71$\pm$2.1 & 83.94$\pm$1.9 & 85.42$\pm$2.0  \\
\hline
ResNet-101 &88.14$\pm$2.1  & 89.02$\pm$1.8 & 88.43$\pm$1.6 & 90.63$\pm$1.5  \\
\hline
Inception &89.33$\pm$2.5  & 89.73$\pm$2.2 & 88.91$\pm$2.1 & 90.89$\pm$2.1  \\
\hline
Inc.-ResNet-v2 & 87.34$\pm$2.3 & 87.44$\pm$2.4 & 86.72$\pm$2.3 & 88.75$\pm$2.2   \\
\hline
MobileNetV2 & 91.24$\pm$2.4  & 92.44$\pm$2.3 & 88.09$\pm$2.1 & 91.75$\pm$2.2  \\
\hline
\textbf{Mobincep} &\textbf{94.82$\pm$2.1} & \textbf{94.86$\pm$1.9} & \textbf{94.11$\pm$1.8} & \textbf{95.90$\pm$1.8}  \\
\Xhline{5\arrayrulewidth}
\end{tabular}
\end{adjustbox}
\end{table}

\renewcommand{\arraystretch}{1.1}
\begin{table*}
\centering

\caption{Classification metrics per class for separate (at left) and mixed (at right) datasets.}
\label{table_accuracy_metrics}
\centering
\begin{adjustbox}{width=1\textwidth}

\begin{tabular}{l|c|c|c|l|c|c|c}
\Xhline{5\arrayrulewidth}
\textbf{Dataset: Class} & \textbf{Precision} & \textbf{Sensitivity} & \textbf{F1} & \textbf{Dataset: Class} & \textbf{Precision} & \textbf{Sensitivity} & \textbf{F1}\\
\Xhline{2\arrayrulewidth}
Pap: Normal	& 0.888 & 0.882 & 0.885 &  Mix: Normal & 0.906 & 0.874 & 0.890
 \\ 
Pap: Abnormal & 0.956 & 0.960 & 0.958 &  Mix: Abnormal & 0.954 & 0.966 & 0.960
 \\ 
 \hline
Lym: MCL & 0.952 & 0.886 & 0.918 &  Mix: MCL & 0.966 & 0.888 & 0.925
 \\ 
Lym: FL & 0.952 & 0.984 & 0.968 &  Mix: FL & 0.940 & 0.962 & 0.951
 \\ 
Lym: CLL & 0.926 & 0.946 & 0.936 &  Mix: CLL & 0.908 & 0.956 & 0.931
 \\ 
  \hline
Hela: Nucleolus & 0.988 & 1.000 & 0.994 &  Mix: Nucleolus & 0.976 & 0.988 & 0.982
 \\ 
Hela: Mitochondria & 0.956 & 0.894 & 0.924 &  Mix: Mitochondria & 0.904 & 0.948 & 0.925
 \\ 
Hela: Microtubules & 0.952 & 0.988 & 0.970 &  Mix: Microtubules & 0.966 & 0.954 & 0.960
 \\ 
Hela: Lysosome & 0.944 & 0.928 & 0.936 &  Mix: Lysosome & 0.928 & 0.940 & 0.934
 \\ 
Hela: Golgpp & 0.946 & 0.904 & 0.925 &  Mix: Golgpp & 0.974 & 0.846 & 0.905
 \\ 
Hela: Golgia & 0.912 & 0.940 & 0.926 &  Mix: Golgia & 0.876 & 0.966 & 0.919
 \\ 
Hela: ER & 0.958 & 0.952 & 0.955 &  Mix: ER & 0.932 & 0.964 & 0.948
 \\ 
Hela: Endosome & 0.928 & 0.914 & 0.921 &  Mix: Endosome & 0.926 & 0.838 & 0.880
 \\ 
Hela: DNA & 0.978 & 1.000 & 0.989 &  Mix: DNA & 0.978 & 1.000 & 0.989
 \\ 
Hela: Actin & 1.000 & 1.000 & 1.000 &  Mix: Actin & 0.990 & 1.000 & 0.995
 \\

\Xhline{5\arrayrulewidth}
\end{tabular}
\end{adjustbox}
\end{table*}

\subsection{Analysis of classification results}
\label{result_analysis}

We compared the classification accuracy of our approach to recent deep CNN models pre-trained on the ImageNet dataset: VGG19  \cite{VGG2015}, GoogleNet \cite{GoogleNet2015}, ResNet-101 \cite{Resnet2016}, Inception \cite{Inception2016}, and Inception-ResNet-v2 \cite{SzegedyIV16}. We also trained the lightweight CNN model MobileNetV2 \cite{Mobilenetv22018} from scratch using the same training conditions as in the original research. This model has very low computation and model complexity and thus we could train it from scratch with our limited training data. 

As shown in Table \ref{table_SOA_1}, our approach achieves better top-1 accuracy than the pre-trained deep CNN models either when learning images from three domains simultaneously in the Mix dataset or when learning from each of the single domain datasets. Notably, it gains over 10$\%$ accuracy compared to the VGG or GoogleNet network. Compared to the trained-from-scratch MobileNetV2, our model also produces better top-1 accuracy across the four datasets. More importantly, our performance results are more consistent in the Lymphoma dataset.  

\begin{figure*}
	\centering	\includegraphics[width=\textwidth]{ 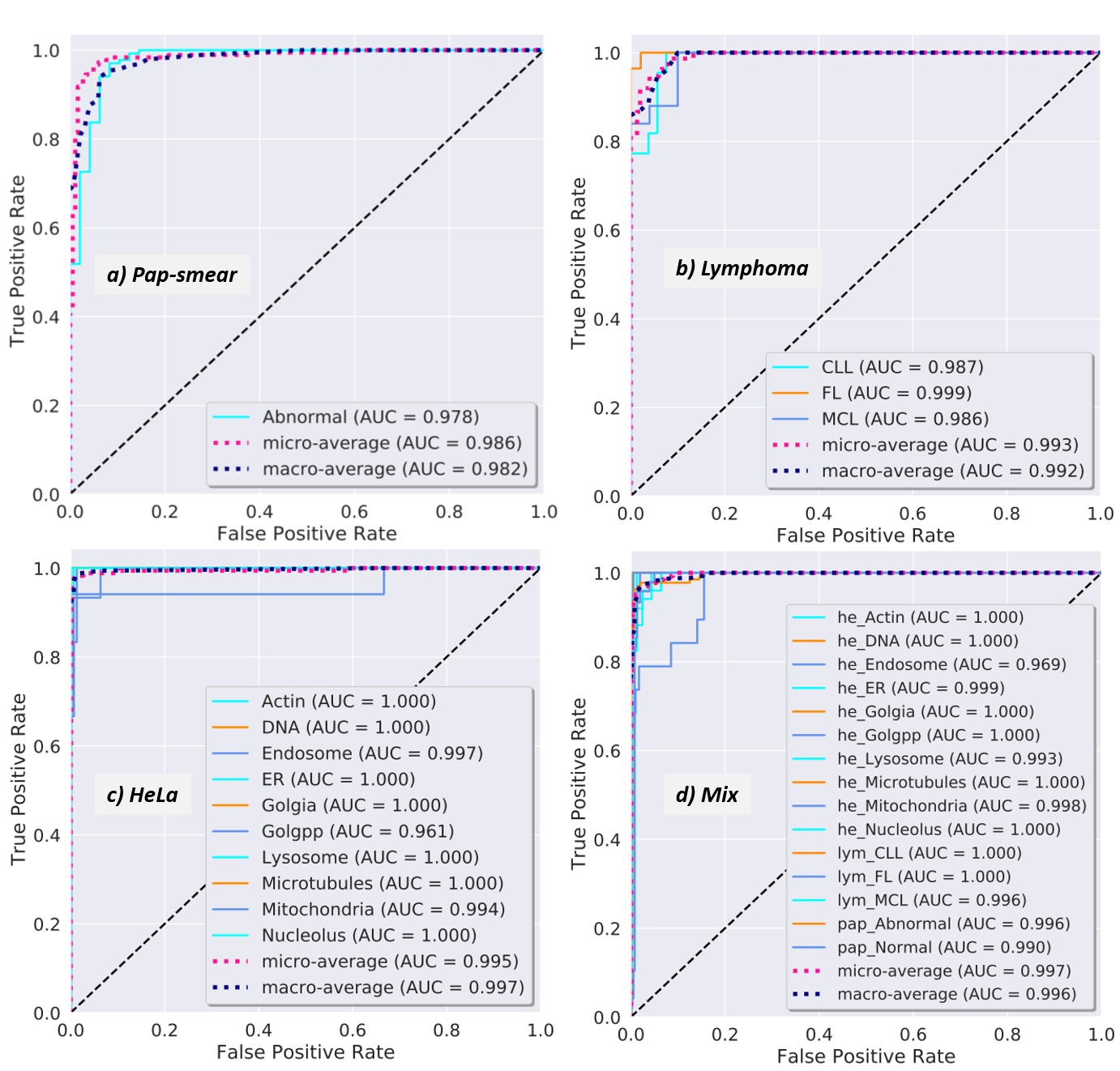}
\caption{ROC curves for the classification task on the different datasets: a) Pap-smear, b) Lymphoma, c) HeLa, d) Mix. The micro-average AUC score is calculated sample-wise, computing average value across all classes weighted by the number of samples in each class, whereas the macro-average score is class-wise, computing unweighted average value across classes.}
\label{Fig:ROCs}
\end{figure*}

We examined the classification metrics of the proposed model in the SDL and MDL modes, as shown in Table \ref{table_accuracy_metrics}. When learning on the mixed dataset, the precision and sensitivity of detecting abnormal samples were $95.4\%$ and $96.6\%$, respectively, while the F1 score was $96.0\%$. These scores are slightly better than when the network was trained on the Pap-smear dataset alone. Figure \ref{Fig:ROCs}a and \ref{Fig:ROCs}d plots the true-positive rates (TPR) against the false-positive rates (FPR), known as ROC curves, for classifying abnormal samples from normal samples in the SDL and MDL modes, respectively. In the case of MDL, the AUC value is $0.996$, which is also better than the single-domain value ($0.978$). 

For the Lymphoma images, the F1 score values were higher than $92.2\%$ for all three classes. We can also notice a significant variation in the precision and sensitivity across the classes, e.g. the low sensitivity for MCL compared with FL and CLL. This fluctuation is similar in both MDL and SDL. In practice, we could improve the classification performance by using a suitable decision threshold. This value could be selected based on the ROC curve for the MCL class in Figure \ref{Fig:ROCs}b, where the AUC of the MCL class is close to the CLL and FL classes.

For the HeLa dataset, the MDL model yielded F1 scores of over $90.4\%$, except for the Endosome class which has a score of $87.6\%$. Again, this lower performance resembles the network’s output when learning from the single dataset. These results coincide with the fact that experts find it challenging to distinguish between Endosomes and Lysosomes or between Golgpp and Golgia proteins. 
The ROC curves in Figure \ref{Fig:ROCs}d confirm that the proposed classifier works well in recognizing subcellular organelles in fluorescent images as the AUC values are at least $0.969$. 

As can also be seen from the Figure \ref{Fig:ROCs}, we obtained micro-average and macro-average AUC scores close to 1.0 in all datasets.

\subsection{Impact of available training data}
\label{training_ratio_analysis}
\begin{figure*}
	\centering	\includegraphics[width = 0.5\textwidth]{ 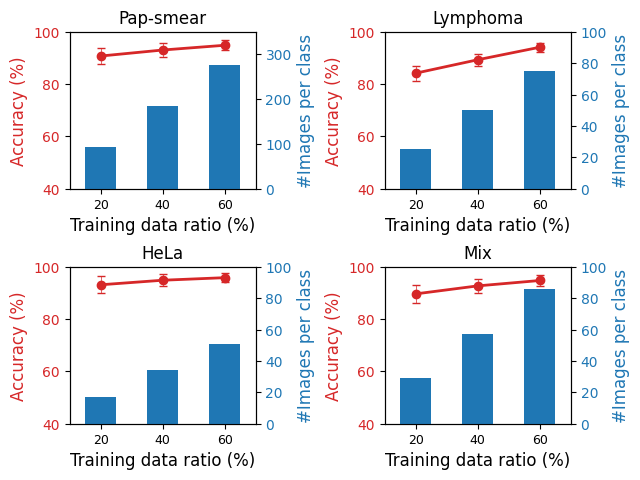}
\caption{Impact of training data volume on classification accuracy. }
\label{Fig:data_ratio_impact}

\end{figure*}

Next, we investigated the impact of limiting the number of labeled images on the model’s performance. For each dataset, we experimented with three different ratios for 5-fold cross-validation by increasing the ratio of training data: 20/20/60, 40/20/40, and 60/20/20. We illustrate the results in Figure \ref{Fig:data_ratio_impact}. 

For the Pap smear dataset, it needs to learn from around 90 training images per class to distinguish between normal and abnormal classes with an accuracy of $90.74\%$. This accuracy level increased to $94.86\%$ when the number of images available for training tripled. On the other hand, the classifier required only about 50 and 20 images per class to reach an accuracy of around $90\%$ on the Lymphoma and HeLa datasets, respectively. 
By comparison, existing machine learning methods require at least 70 labeled images per class to achieve cross-validation accuracy close to $90\%$ on fluorescence microscopy images like those in the HeLa dataset \cite{YuISBI2009}. 
In MDL mode on the Mix dataset, we can see that Mobincep reached the $90\%$ accuracy level with 30 images per class.
These results not only attest to the classifier’s generalization ability on unseen data but also reveal a very useful property of Mobincep, which is to reduce the costly labeling effort needed from experts.

\subsection{Impact of the regularization technique}
\label{reg_feature_analysis}
To validate the contribution of the proposed regularization technique, we compared the performance of the network with the baseline case where only the conventional cross-entropy loss function (the $L_{CE}$ term in equation (\ref{eqn:cost_function})) was used during optimization.

As shown in Table \ref{table_strategy_analysis}, for all experimented datasets, regulating features during optimization improved the classification accuracy. On the multi-domain Mix dataset, it allowed the model to gain $1.56\%$ in average accuracy.\\

\renewcommand{\arraystretch}{1.6}
\begin{table}
\centering
\huge
\caption{Impact of feature regularization on classification accuracy ($\%$)}
\label{table_strategy_analysis}
\begin{adjustbox}{width=0.48\textwidth}
\begin{tabular}{ l|c|c|c|c }
\Xhline{5\arrayrulewidth}
\textbf{Model}  & \textbf{Mix} & \textbf{Pap} & \textbf{Lym} & \textbf{HeLa}\\
\Xhline{5\arrayrulewidth}
Baseline & 93.26$\pm$2.6 & 93.08$\pm$2.4 & 91.23$\pm$2.2 & 93.15$\pm$2.1   \\
\hline
\textbf{Mobincep}&\textbf{94.82$\pm$2.1} & \textbf{94.86$\pm$1.9} & \textbf{94.11$\pm$1.8} & \textbf{95.90$\pm$1.8}  \\
\Xhline{5\arrayrulewidth}
\end{tabular}
\end{adjustbox}
\end{table}

\subsection{Comparison with state of the art methods}
\label{SOAcompare}
In Table \ref{table_SOA_2}, we compare our proposed Mobincep network with recent methods that achieved the highest published results on the three experimented datasets (Pap, Lym, and Hela).

Looking at the Pap smear dataset, all of the methods using a single deep CNN network give no improvement over conventional hand-crafted feature-based methods, reaching accuracy levels lower than $91\%$. The accuracy increases notably only when multiple deep networks are combined, reaching around $93\%$ as achieved in \cite{long2018}. In contrast, our approach produces the best performance by training only a lightweight CNN model with much lower complexity, with $94.02\%$ accuracy in MDL mode and $94.86\%$ in SDL mode.

The Lymphoma dataset appears to be more challenging for designing suitable hand-crafted feature descriptors. The best approach which was proposed in \cite{nanni2020}, with $93.87\%$ accuracy, merged 8 different deep CNN models. However, we show that while our Mobincep network has only $4.3$M parameters, it produces an accuracy of more than $94\%$. 

For the HeLa dataset, the Capsule Neural Network is the best option among CNN models, but even its accuracy of $93.08\%$ is still well below the top performance ($95.3\%$) produced by the hand-crafted method described in \cite{chebira2007}. Our Mobincep network could surpass this value, reaching $95.9\%$ accuracy whether training on the mixed dataset or the HeLa dataset alone. This confirms yet again the effectiveness of our model and its ability to generalize across different image domains. 

In the multi-domain learning setting, the proposed network achieves classification accuracy roughly equivalent to its performance when it is optimized on each domain separately. Indeed, while its accuracy decreased by $0.84\%$ on the Pap-smear classes, the average accuracy was maintained on the HeLa classes and slightly improved on the Lymphoma images.

\renewcommand{\arraystretch}{1.0}
\begin{table*}
\caption{Comparison of Mobincep with competing methods on different datasets. All values are accuracies (\%). Those for other methods are as published in the literature, with standard deviation when available. Top-3 values in each dataset are formatted in bold.}
\label{table_SOA_2}
\centering

\begin{tabular}{ l|c|c|c}
\Xhline{5\arrayrulewidth}
\textbf{Model} & \textbf{Pap} & \textbf{Lym} & \textbf{HeLa} \\
\Xhline{3\arrayrulewidth}
Spatial adjacent histogram based on   & $88.03\pm1.7$ & & $	90.06 \pm1.5$\\ 
adaptive local binary patterns+SVM \cite{liu2016}    &   & & \\ 
       \hline
\makecell[l]{SVM cascaded with a reject option and }
     &$90.96 \pm 0.5$ & &$92.96 \pm 1.3$ \\ 
\makecell[l]{subspace analysis \cite{lin2018Biomedical}}
     &  & &  \\      
     
     \hline
    WND-CHARM based on 1025  & & 85.00 & $ 87.00 \pm 9.00$ \\
content descriptors \cite{CPCharm2016, WNDChrm2008} & &   &   \\
    \hline
CP-CHARM based on 953  & & $66.00 \pm 1.0$  & $ 84.00 \pm 0.4$\\
content descriptors \cite{CPCharm2016}  & &    &  \\
        \hline
Fusion of multiple handcrafted and  & &90.67  \\ 
deep learned features \cite{NanniTrans2018} & &   \\ 
     \hline
Multiresolution classification system \cite{chebira2007} & & & \textbf{~95.30}  \\
\Xhline{3\arrayrulewidth}
Pretrained ResNet-101 \cite{nanni2020}  & &86.40  \\ 
        \hline
Pretrained ResNet-152 \cite{long2018} & $90.87 \pm 1.5$ &\\ 
    \hline
Pretrained Inception-ResNet-v2 \cite{long2018} & $89.25 \pm 2.2$ &  & 92.00 \\ 
     \hline
Pretrained Inception-v3 \cite{long2018,nanni2020} & $89.66 \pm1.9$ &87.47 \\ 
     \hline
\makecell[l]{Ensemble of pretrained Inception-v3 \\ and ResNet-152 \cite{long2018}} & $92.38 \pm 1.3$ &\\ 
     \hline
\makecell[l]{Ensemble of pretrained Inception-v3, \\ ResNet-152, Inception-ResNet-v2 \cite{long2018}}  & \textbf{93.04 $\pm$ 1.5}  & & 92.57\\ 
     \hline
\makecell[l]{Fusion of 8 different deep CNN models \cite{nanni2020}}   & & \textbf{93.87} \\
        \hline
GoogleNet \cite{godinez2017}   & & & 92.00  \\ 
        \hline
Capsule Neural Network (CapsNet) \cite{zhang2018Capsule}& & & 93.08 \\ 
        \hline
Multi-scale CNN \cite{godinez2017} & & & 91.00 \\ 

        \hline
\textbf{Mobincep}     & \textbf{94.86$\pm$1.9} & \textbf{~94.11$\pm$1.8} & \textbf{~95.90$\pm$1.8}\\
        \hline
\textbf{Mobincep }     & \textbf{94.02$\pm$1.9} & \textbf{~94.20$\pm$1.8} & \textbf{~95.90$\pm$1.8}\\
\textbf{(Multi-domain learning on Mix dataset)}     &   &  &  \\
\Xhline{5\arrayrulewidth}
\end{tabular}
\end{table*}

\section{Discussion}
This study aimed to develop a multi-domain learning model for the classification of microscopy images from different domains. Our principal contribution is the design of a compact CNN model that can be trained from scratch on target domains that have a very limited number of image samples. The proposed multi-domain learning approach can facilitate the choice of an analysis tool and the configuration of its parameters to account for different microscopes, objects of interest, and imaging conditions. This can accelerate the pace of investigation and reduce the required expertise in adapting computer algorithms. From a design perspective, this limits the need to select domain-specific hyper-parameters and eases the training process, as we can train only once the model and run on different experimental domains.

Prior studies typically used very deep CNNs or ensembles of these architectures to achieve high performance in clinical or biomedical applications. Nevertheless, this increases the requirement for computer hardware that is not always readily available. More importantly, their high complexity generally renders such networks selective for a certain image domain. As described in section \ref{result_analysis}, the very good classification statistics and high AUC values ($>0.95$) on multiple domains demonstrate the benefit of using the Mobincep model as an automatic classifier. Our experiments show that the model yields state-of-the-art performance when learning from multiple microscopy image domains. This can be explained by the fact that the model has very low complexity, thus we can train it from scratch directly on the target data, instead of fine-tuning a pre-trained feature extractor using transfer learning.

\section{Conclusion}
In this work, we presented a lightweight CNN classifier for learning multiple domains of microscopy images. Moreover, we formulated a new optimization function and devised a suitable training strategy allowing our network to outperform state-of-the-art methods. The proposed model performs well in multiple applications of microscopy image classification. Because of its low complexity, the approach becomes more appealing for deployment in clinical and biomedical studies. Further research will focus on developing a domain generalization algorithm, such that the model can work well on new microscopy images that are captured under imaging conditions different from those of training images. 

\section*{Acknowledgments}
The authors would like to thank Philippe Debanné for revising this manuscript.

\bibliographystyle{unsrt}  
\bibliography{references}

\end{document}